\documentclass{article}

\usepackage{PRIMEarxiv}
\usepackage{amsmath} 
\usepackage[utf8]{inputenc} 
\usepackage[T1]{fontenc}    
\usepackage{hyperref}       
\usepackage{url}            
\usepackage{booktabs}       
\usepackage{amsfonts}       
\usepackage{nicefrac}       
\usepackage{microtype}      
\usepackage{lipsum}
\usepackage{fancyhdr}       
\usepackage{graphicx}       
\graphicspath{{media/}}     

\title{SocFedGPT: Federated GPT-based Adaptive Content Filtering System Leveraging User Interactions in Social Networks
}

\author{
  Sai Puppala\\
  School of Computing \\
  Southern Illinois University Carbondale, IL, USA, 62901\\
  \texttt{saimaniteja.puppala@siu.edu} 
  \\
\And
  Ismail Hossain, Md Jahangir Alam, Sajedul Talukder \\
  Computer Science \\
  University of Texas at El Paso, TX, USA, 79968\\
  \texttt{\{ihossain, malam10\}@miners.utep.edu}\\
  \texttt{stalukder@utep.edu}
}

\begin{document}
\maketitle              
\begin{abstract}
Our study presents a multifaceted approach to enhancing user interaction and content relevance in social media platforms through a federated learning framework. We introduce personalized GPT and Context-based Social Media LLM models, utilizing federated learning for privacy and security. Four client entities receive a base GPT-2 model and locally collected social media data, with federated aggregation ensuring up-to-date model maintenance. Subsequent modules focus on categorizing user posts, computing user persona scores, and identifying relevant posts from friends' lists. A quantifying social engagement approach, coupled with matrix factorization techniques, facilitates personalized content suggestions in real-time. An adaptive feedback loop and readability score algorithm also enhance the quality and relevance of content presented to users. Our system offers a comprehensive solution to content filtering and recommendation, fostering a tailored and engaging social media experience while safeguarding user privacy.

\end{abstract}
\section{Introduction}
In an era dominated by the pervasive influence of social media, the ability to sift through vast streams of content and deliver personalized, relevant experiences to users has become paramount \cite{liang2006personalized}. Our research addresses this issue by combining advanced machine-learning techniques with privacy-preserving methodologies. We employ federated learning \cite{zhang2021survey}, a method that allows collaborative model training across distributed devices while safeguarding sensitive user data \cite{hossain2023collaborative}. This innovative approach aims to revolutionize content filtering and recommendation systems \cite{may2014filter}, enhancing user engagement.

Central to our research is the development of context-aware models \cite{zhu2013context} specifically for social media. Using state-of-the-art language models like GPT \cite{floridi2020gpt}, we create personalized frameworks that accurately understand and categorize user-generated content. Beyond categorization, we profile user personas to understand their preferences deeply. By analyzing user interactions and content engagement, we aim to provide highly personalized recommendations that resonate on an individual level.

Existing research has explored various methodologies to address content filtering and recommendation in social media ecosystems \cite{may2014filter}. Using advanced language models like GPT, context-aware modeling techniques have shown promise in understanding user-generated content \cite{titus2024does}. By leveraging a GPT-like decoder-only model, user evolution in social media can be implemented with the understanding of user networks, user demographics, user history, and user engagement~\cite{hossain2024evolve}. These models categorize posts and discern user preferences, forming the basis for personalized content delivery. Persona profiling techniques \cite{li2022persona} capture diverse user interests and behaviors, enhancing content relevance \cite{chaoji2012recommendations}.

Despite these advancements, challenges remain in balancing personalized experiences \cite{neuhofer2015smart} with user privacy. Federated learning offers a solution by decentralizing model training and keeping data localized. Combining federated learning with context-aware and persona-based modeling, our research aims to advance content filtering and recommendation systems, creating a more personalized, secure, and engaging user experience.

Building upon previous research \cite{puppala2022towards}\cite{talukder2022novel}, our approach integrates federated learning with context-aware and persona-based modeling techniques to tackle content filtering and recommendation challenges in social media. Using federated learning \cite{talukder2022federated}, we emphasize user privacy and data security while enabling collaborative model training across distributed client devices. This decentralized method maintains data integrity and promotes inclusive model development.

Our methodology employs personalized GPT \cite{kim2023towards} and Context-based Social Media LLM models to categorize content and profile user personas. These models are trained locally on client devices using federated learning, ensuring privacy and their aggregated results update a global model that captures diverse user interests. Our research aims to enhance content experiences and set new privacy standards in social media analytics by combining federated learning with context-aware modeling and persona profiling. Our experiments demonstrate the approach's effectiveness and scalability, advocating for the coexistence of personalized content and data privacy in social media.

The paper is structured to provide an optimizing content filtering and recommendation systems in social media using federated learning. The methodology section details our technical approach, including model deployment, federated learning techniques, and context-aware and persona-based modeling. Results demonstrate the approach's efficacy and scalability, followed by a discussion of key findings, limitations, and future directions. The conclusion summarizes key insights, emphasizes user privacy, and highlights the practical implications of our work.

\section{Methodologies}
\subsection{Problem statement}
Given the increasing reliance on social media platforms for content consumption and interaction, there is a pressing need for advanced content filtering and recommendation systems that can deliver personalized experiences while ensuring user privacy. Traditional centralized approaches to content filtering may compromise individual data integrity and struggle to adapt to the dynamic nature of user preferences across different social media platforms. To address these challenges, our research aims to develop a decentralized content filtering and recommendation framework utilizing federated learning techniques. By deploying personalized context-aware models tailored for social media content categorization and user persona profiling, our goal is to enable collaborative model training across distributed client devices while preserving user privacy. Through this approach, we seek to optimize content filtering and recommendation systems, offering users personalized and relevant content experiences without compromising their privacy.

\begin{figure}
    \centering
    \includegraphics[width=\textwidth]{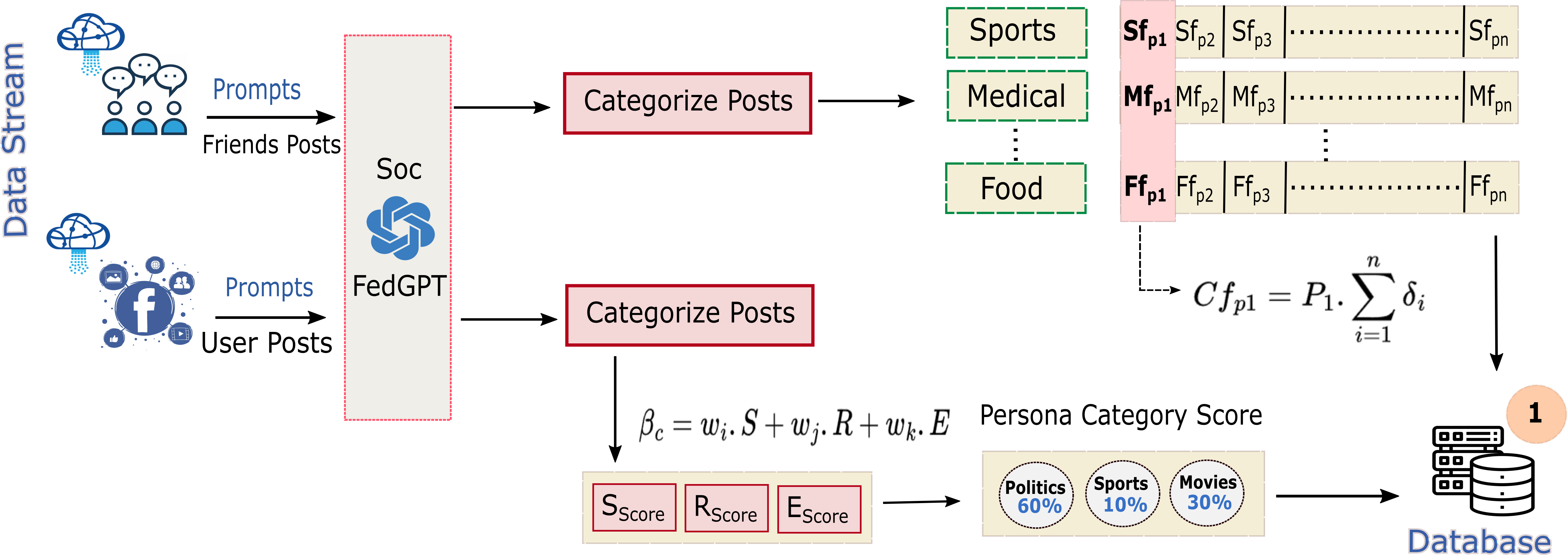}
    \caption{The system architecture elucidates various components in the Adaptive Content Filtering System approach, with a key focus on GPT, User Persona profile, and Category-based social engagement. Additionally, it provides a concise overview of the distinctive features of the federated learning global server for context-based GPT generation. }
    \label{fig:figure2-overtime}
\end{figure}

\subsection{Federated Learning}
Federated learning is a decentralized method for training models while maintaining data privacy, as client data remains on their devices. Our approach involves training the GPT model across multiple clients, who share only model updates, not raw data.

Each client starts with common model parameters and trains the GPT model using their local data, resulting in unique model updates. These updates are aggregated to form a global model that incorporates insights from all clients, without exchanging raw data. This iterative process periodically updates the global model to reflect the combined knowledge of all clients. This method preserves data privacy and generates a more personalized GPT model by leveraging diverse data from multiple clients. The equations illustrate how global and local model parameters are updated and aggregated to improve the model iteratively.

\begin{equation}
\begin{aligned}
\text{Global Parameter Update:} & \quad \phi^{(t)} \leftarrow \phi^{(t-1)} - \alpha \nabla J(\phi^{(t-1)}, D^{(t)}) \\
\text{Client Update:} & \quad \phi_{i}^{(t)} \leftarrow \phi^{(t)} - \alpha \nabla J_i(\phi^{(t)}, D_i) \\
\text{Aggregation:} & \quad \phi^{(t+1)} \leftarrow \frac{1}{N} \sum\limits_{i=1}^{N} \phi_{i}^{(t)}
\end{aligned}
\end{equation}

$\phi^{(t)}$ represents the global model parameters at iteration $t$.
$\alpha$ is the learning rate.
$\nabla J(\phi^{(t-1)}, D^{(t)})$ denotes the gradient of the loss function $J$ with respect to the global model parameters using the local dataset $D^{(t)}$.
$\phi_{i}^{(t)}$ represents the local model parameters of client $i$ at iteration $t$.
$\nabla J_i(\phi^{(t)}, D_i)$ denotes the gradient of the loss function $J$ with respect to the local model parameters of client $i$ using its local dataset $D_i$.
The aggregation step computes the updated global model parameters by averaging the local model parameters from all participating clients.

\subsection{User Persona Profiler}
User interaction data, including likes, shares, and comments, is collected from the LLM platform. Each post is categorized by content (e.g., sports, politics) and sentiment (positive, negative, neutral). Data prepossessing ensures accurate categorization and sentiment labeling by cleaning and standardizing the data. We analyze user engagement within each content category to determine preference distributions and sentiment distributions (positive, negative, neutral). Interactions are weighted~\cite{hossain2024socialrec} to reflect their significance in user preferences and sentiments.

To understand user personas, we compute a composite \textit{Content Score} based on Engagement, Readability, and Sentiment. This score measures content quality and audience resonance, excluding Originality Score due to the difficulty in quantifying uniqueness in large social media datasets.
\subsubsection{Engagement Score ($E$)}

The Engagement Score encapsulates user interactions with the content, including likes, shares, comments, and views. Each interaction type is assigned a distinct weight based on its perceived value in denoting user engagement. The Engagement Score is normalized to a $[0,1]$ scale and calculated as follows:

\begin{equation} \label{eq:engagement_score}
\begin{split}
E = & \frac{w_{\text{likes}} \cdot \text{Likes} + w_{\text{shares}} \cdot \text{Shares} }{MaxE} \\
& + \frac{w_{\text{comments}} \cdot \text{Comments}}{MaxE}
\end{split}
\end{equation}

where $w_{\text{likes}}$, $w_{\text{shares}}$, and $w_{\text{comments}}$ are the weights assigned to likes, shares, comments, and views, respectively, and $MaxE$ is the maximum possible engagement score for normalization purposes.

\subsubsection{Sentiment Score ($S$)}
The Sentiment Score reflects the overall sentiment of posts and comment threads on social media. The score is normalized to the $[0,1]$ range, where $0$ represents entirely negative sentiment, $0.5$ represents neutral sentiment, and $1$ represents entirely positive sentiment.

\subsubsection{Category Readability Score ($\rho$)}
In the context of assessing user engagement through readability, we developed a Readability Score metric that categorizes posts based on the complexity and clarity of their language. This metric assigns a score of `1' to posts written with simple language, easily accessible to a general audience, enhancing widespread comprehension and interaction. Posts that utilize professional, industry-specific terminology are assigned a higher score of `2', reflecting their targeted appeal to a professional demographic that values technical accuracy or specialized knowledge. Conversely, posts deemed unreadable, either due to poor grammar, lack of coherence, or excessive complexity without proper context, receive a score of `0'. 

This scoring approach underscores the importance of clarity and audience-tailored communication in written content. To compute an aggregate Readability Score for each user, we average the scores of all posts they have interacted with, providing a quantifiable measure of the typical readability level of content that the user engages with. This methodology allows for a nuanced analysis of user preferences and content interaction patterns, which is crucial for tailoring content strategy and improving user engagement.

\subsubsection{User Persona Score ($C$)}

The Persona Score ($C_k$) (Here ($k$) represents a specific category) is a weighted sum of the Engagement ($E$), Readability($R$), and Sentiment ($S$) scores, each contributing to the final score based on predefined weights that reflect their relative importance. The formula for calculating $C$ is as follows:

\[ \beta_k = w_{i} \times E + w_{j} \times S + w_{k} \times \rho \]

where $w_E$, $w_R$, and $w_S$ are the weights for the Engagement, Readability, and Sentiment scores, respectively, satisfying $w_E  + w_S + w_R = 1$.

The comprehensive framework would look similar to below,
\[
\beta_{distribution} = \begin{bmatrix}
\beta_{\text{news}} / (\beta_{\text{news}} + \beta_{\text{media}} + \beta_{\text{politics}} + \cdots) \\
\beta_{\text{media}} / (\beta_{\text{news}} + \beta_{\text{media}} + \beta_{\text{politics}} + \cdots) \\
\beta_{\text{politics}} / (\beta_{\text{news}} + \beta_{\text{media}} + \beta_{\text{politics}} + \cdots) \\
\vdots
\end{bmatrix}
\]

This methodology provides a comprehensive framework for evaluating online content, leveraging quantitative metrics to assess user engagement, content accessibility, and emotional resonance. The resultant Content Score serves as a versatile tool to gauge content quality and effectiveness in digital communication for content filtering.

\begin{figure}
\centering
\includegraphics[width=0.8\textwidth]{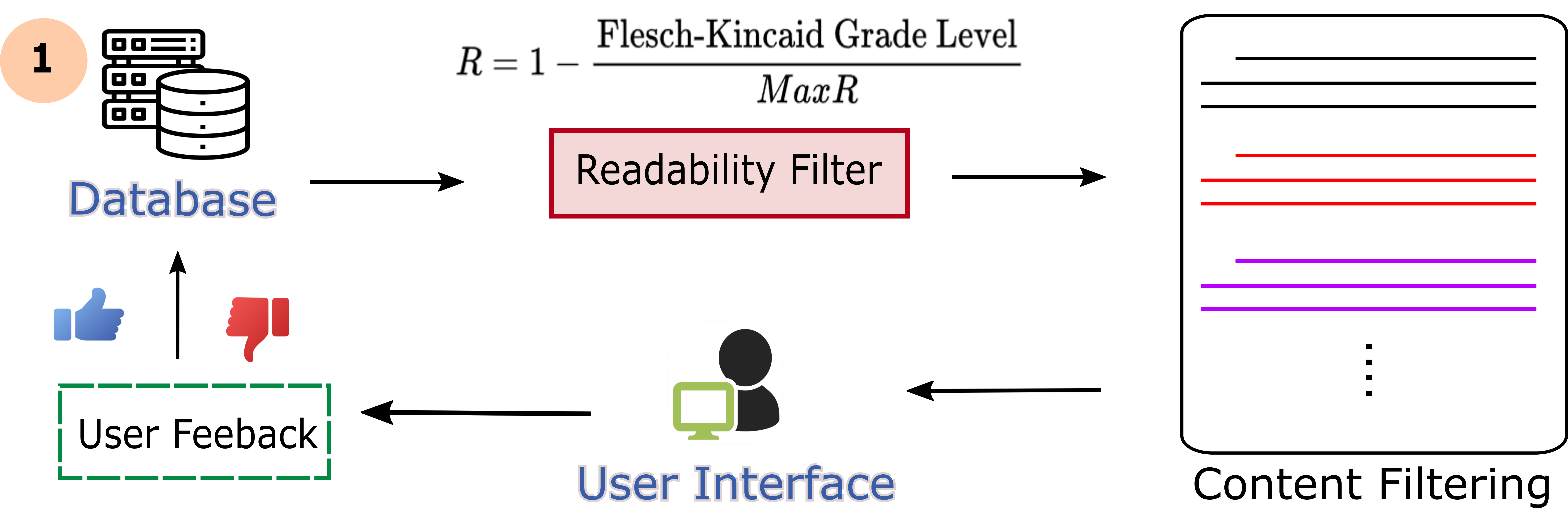}
\caption{Continuation of figure 1: The workflow architecture of adaptive content filtering starting from database and all the way to feedback collection from user.}
\label{fig: figure-overtime}
\end{figure}

\subsection{Content Filtering}
In our approach to content filtering, we prioritize the personalization of user experience by initially computing a `user rank' for each individual within a user's social circle. This rank is derived from a carefully formulated scoring system, which quantifies the user's engagement with content shared by each friend. Specifically, the score for each friend is calculated using the following formula:

\[ {{\delta}_{i}} = w_l \times L_i + w_c \times C_i + w_{sh} \times Sh_i \]

Here, \(\delta_i\), \( L_i \), \( C_i \), and \( Sh_i \) represent the overall friend i engagement score, number of likes, comments, and shares, respectively, that the current user has made on the content posted by friend \( i \). The weights \( w_l \), \( w_c \), and \( w_{sh} \) correspond to the relative importance of likes, comments, and shares in determining the strength of engagement and affinity towards that friend. 

This scoring mechanism allows us to assess the degree of interaction and preference the user exhibits towards each friend's content. By leveraging this data, our system can more accurately filter and prioritize content that aligns with the user's demonstrated interests and interactions within their network. This methodology not only enhances content relevance but also improves user satisfaction by curating a feed that reflects personal connections and preferences.

upon calculating specific scores for each user derived from their engagement levels, we proceed to tailor the content delivery mechanism. This customization involves filtering categorical posts that align with the interests identified in the user's persona score calculation. For example, if the persona score indicates a preference for sports, the system filters and prioritizes sports-related posts or articles shared by friends within the user's network, particularly those friends who have demonstrated high engagement levels as reflected in our engagement score matrix. This method ensures that the content presented to the user not only aligns with their interests but also emanates from highly interactive sources within their social circle, thereby enhancing user engagement and content relevance. The below equation formulates the importance of the friend post $P$.

\[
{P_i} = w_C \cdot C + w_L \cdot L + w_S \cdot S + w_T \cdot \frac{1}{T}
\]

Here we employ weighted factors where $w_C$, $w_L$, $w_S$, and $w_T$ represent the weights assigned to comments, likes, shares, and latency of the post, respectively. The variables $C$, $L$, and $S$ denote the number of comments, likes, and shares that a post has received. The variable $T$ indicates the time elapsed since the post was published. In the expression, multiplication is denoted by $\cdot$ to enhance clarity and visual distinction. The term $\frac{1}{T}$, representing the reciprocal of $T$, is used to increase the score of more recent posts, thereby emphasizing the timeliness of content in its valuation. 

\subsubsection{Additional Filtering}

Additionally, our system incorporates an enhanced filtering mechanism that enables users to selectively exclude posts with negative scores, utilizing sentiment analysis and predictive scoring of comment thread trends. Furthermore, we apply another layer of filtering based on categories derived from the user's persona profile, allowing for content to be tailored more closely to individual preferences. Posts that do not fit into these specified categories are classified as general posts, providing a broad spectrum of content while still prioritizing personalized user interests.
For each post \(i\), we apply a filter based on sentiment and trend predictions:
\[
F_i = \begin{cases} 
1 & \text{if } S_i > 0 \text{ and } T_i > \tau \\
0 & \text{otherwise}
\end{cases}
\]
Here, \(\tau\) represents a threshold for the trend prediction score that determines if a post is trending positively and thus interesting to the user. \(F_i\) represents the filtering status of post \(i\), where 1 indicates the post passes the filter and 0 indicates it does not.

\subsubsection{Readability Score ($R$)}

The Readability Score assesses the ease with which an audience can comprehend the content. It is determined using the Flesch-Kincaid readability test, which is based on the length of words and sentences. The score is normalized to a $[0,1]$ scale, with higher scores indicating more accessible content:

\[ R = 1 - \frac{\text{Flesch-Kincaid Grade Level}}{MaxR} \]

where $MaxR$ is the maximum grade level used for normalization, ensuring $R$ falls within the $[0,1]$ range. \newline 

\subsubsection{Adaptive Feedback Loop}
To enhance flexibility in our system, we introduced an additional field for user feedback. For instance, a user who previously expressed interest in Community Services may receive suggestions related to this category based on our historical score calculations. However, recognizing that user preferences can change over time as individuals evolve, we implemented a feature allowing users to indicate their preferences using like or dislike buttons. This user feedback is then utilized to update our database records and serve as filtering criteria for future recommendations, ensuring that users receive more relevant and tailored suggestions.

\section{Experiment}
\subsection{Context GPT}
Our system implementation includes different modules, starting with creating a personalized GPT model using federated learning for privacy and security. We connect four client entities to a global server. Each client, equipped with a base GPT-2 model and local social media data collected via web crawlers, trains the model and sends updates back to the server. The global server performs federated aggregation to average these models and redistributes the updated model to clients for further training. This continuous iterative process ensures the model stays current with new data, enhancing adaptability and relevance for our content filtering system.

As depicted in Figure~\ref{fig:figure2-overtime}, the Context-based GPT model, tailored for social media content filtering, plays a pivotal role in categorizing user posts. The stream of user posts is directed to the Context GPT model to ascertain their appropriate categories. To enhance the accuracy of category determination, we employ prompt engineering techniques such as zero-shot and few-shot prompt engineering \cite{yong2023prompt}. Once the categories of the streaming posts are identified based on user history, the subsequent step involves determining the user persona scores for each category. The primary objective of user persona profiling is to discern the user's preferences post-filtering. For example, if user history indicates a preference distribution of 30 percent politics and 70 percent sports, these insights are stored as the ultimate outcome of the user persona in our database, to be utilized as filtering criteria.

\begin{figure}
\centering
\includegraphics[width=\textwidth]{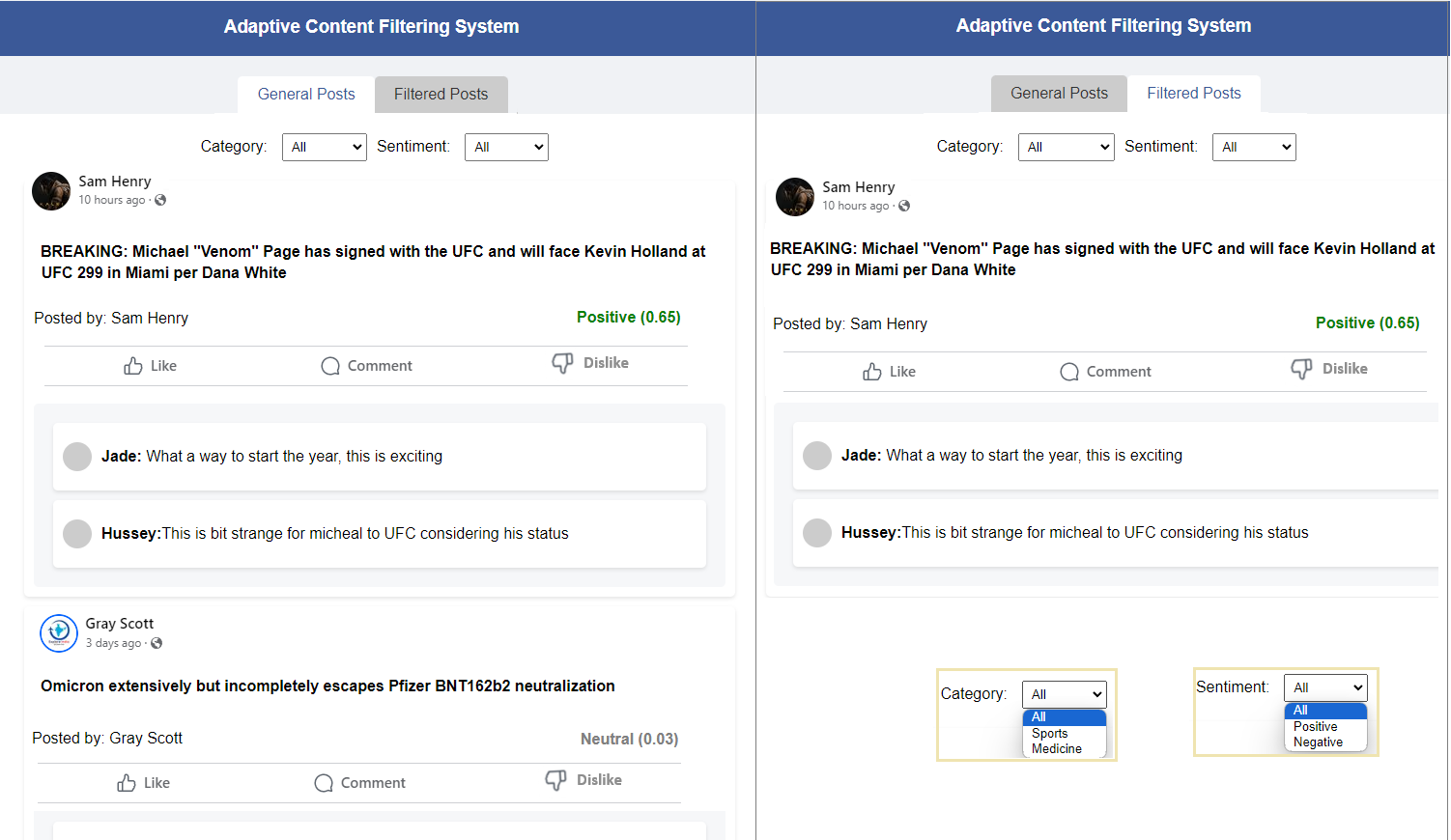}
\caption{User interface for Federated GPT-based Adaptive Content Filtering System Leveraging User Interactions in Social Networks}
\label{fig: figure-overtime}
\end{figure}

\subsection{User Category Persona}
To calculate user persona scores, we analyze user engagement (likes, shares, comments) across different categories from their history. Each category (e.g., politics, sports) is segmented and examined for engagement types. Engagements are assigned weighted scores, with comments, shares, and likes considered. While these weights are currently fixed, they could be adjusted dynamically. Sentiment scores (positive or negative) are also calculated to help filter content. The final user persona score, derived from these analyses, is stored in the database for personalized content suggestions.

\subsection{Content Filtering}
The next module identifies and ranks relevant posts from a user's friends list. It uses social engagement metrics and a context-based GPT model to categorize posts. After categorization, posts are ranked using content filtering techniques based on user preferences. The aim is to suggest related posts to users and identify similar interests within user groups. Interaction scores help determine the most engaging posts for each category.

Posts are selected for user suggestions based on user persona and relevance. The system updates in real time, ensuring constant relevance. Users can filter out negative or positive posts and provide feedback through like/dislike buttons, enhancing personalization. Spam posts are filtered using a readability score algorithm, ensuring high-quality content.

\section{Conclusion}
Our experiment highlights a multi-module system that enhances user experience and content relevance on social media Using personalized GPT models and federated learning, it ensures privacy and adapts to user preferences. The system improves post categorization accuracy, provides tailored content suggestions, and uses social engagement metrics for relevance. Real-time updates and adaptive feedback enhance interaction, while a readability algorithm maintains quality. The system delivers personalized, high-quality content, with potential future improvements in engagement scores and filtering algorithms.

\end{document}